\documentclass[letterpaper, 10 pt, conference]{ieeeconf}  
\IEEEoverridecommandlockouts                              
\usepackage{makecell}
\usepackage{cite}
\usepackage{ulem} 
\usepackage{balance}
\usepackage{graphicx} 
\usepackage{stfloats}
\usepackage{hyperref}   
\usepackage{float}
\usepackage[ruled]{algorithm2e}
\usepackage{caption}
\usepackage{multirow} 
\usepackage{booktabs}
\usepackage{indentfirst}
\usepackage{subfigure}
\usepackage{color}  
\usepackage{lipsum} 
\usepackage{amsmath,esint} 
\usepackage{cuted}
\usepackage{CJKutf8}
\usepackage{verbatim}
\usepackage{bm}
\usepackage{setspace}
\usepackage{array,caption}
\usepackage{fancyhdr}
\usepackage{ulem}
\usepackage{color}

\newcommand{\redcolor}[1]{\textcolor{red}{#1}}

\hypersetup{
	colorlinks=true,
	linkcolor=blue,
	citecolor=blue,
	urlcolor=blue,
}

\title{\LARGE \bf
Design and Validation of an Under-actuated Robotic Finger with Synchronous Tendon Routing
}

\author{Quan Yuan$^\dagger$, Zhenting Du$^\dagger$, Daqian Cao, and Weibang Bai$^\ast$ 
\thanks{
This work is supported by the Shanghai Pujiang Program under grant 23PJ1408500, Shanghai Frontiers Science Center of Human-centered Artificial Intelligence (ShangHAI), MoE Key Laboratory of Intelligent Perception and Human-Machine Collaboration (KLIP-HuMaCo).
The experiments of this work were supported by the Core Facility Platform of Computer Science and Communication, SIST, ShanghaiTech University.}
\thanks{$^\dagger$These authors contributed equally to this work.}
\thanks{$^*$Corresponding author: Weibang Bai \textit{(wbbai@shanghaitech.edu.cn)}.}
\thanks{Quan Yuan, Daqian Cao, and Weibang Bai are with the ShanghaiTech Automation and Robotics (STAR) Center, School of Information Science and Technology, ShanghaiTech University, Shanghai, 201210, China.}
\thanks{Zhenting Du is with King's College London, London, UK; and was with the ShanghaiTech Automation and Robotics (STAR) Center, School of Information Science and Technology, ShanghaiTech University,  Shanghai, 201210, China.}
}

\begin{document}

\maketitle
\thispagestyle{empty}
\pagestyle{empty}

\begin{abstract}
Tendon-driven under-actuated robotic fingers provide advantages for dexterous manipulation through reduced actuator requirements and simplified mechanical design. However, achieving both high load capacity and adaptive compliance in a compact form remains challenging. This paper presents an under-actuated tendon-driven robotic finger (UTRF) featuring a synchronous tendon routing that mechanically couples all joints with fixed angular velocity ratios, enabling the entire finger to be actuated by a single actuator. This approach significantly reduces the number of actuators required in multi-finger hands, resulting in a lighter and more compact structure without sacrificing stiffness or compliance. The kinematic and static models of the finger are derived, incorporating tendon elasticity to predict structural stiffness. A single-finger prototype was fabricated and tested under static loading, showing an average deflection prediction error of 1.0 mm (0.322\% of total finger length) and a measured stiffness of $1.2\times10^{3}$ N/m under a 3 kg tip load. Integration into a five-finger robotic hand (UTRF-RoboHand) demonstrates effective object manipulation across diverse scenarios, confirming that the proposed routing achieves predictable stiffness and reliable grasping performance with a minimal actuator count.




\end{abstract}

\vspace{4mm}
\section{INTRODUCTION}

In recent years, rapid advancements in robotic locomotion and embedded Artificial Intelligence have garnered widespread attention for humanoid robot applications \cite{10415857}. Meanwhile, robotic hands, as the primary interface for interacting with the environment, have emerged as a prominent research focus.
Numerous studies therefore have focused on designing robotic hands with various features, such as high degrees of freedom\cite{dai2009orientation,8794277}, high stiffness\cite{9691882}, fingertip force sensing\cite{10802272}, and excellent compliance\cite{8589671} to enhance the dexterity and compliant manipulation capabilities of robotic fingers. However, practical multi-finger hands often require a large number of actuators to achieve independent joint motion, which increases system weight, volume, wiring complexity, and cost, and limits the compactness of the overall mechanism. Reducing actuator count while maintaining both high stiffness for load-bearing and compliance for adaptive interaction remains a key challenge in compact under-actuated finger design.


Tendon-driven under-actuated robotic fingers offer an affordable solution by reducing the number of actuators and simplifying the mechanical design, providing excellent compliance for dexterous manipulation \cite{7915456}. Tendon-driven mechanisms place actuators outside the fingers to reduce distal inertia and provide natural compliance \cite{dong2018grasp,10312190,9536104}, while under-actuated mechanisms can reduce actuation structural complexity and lower costs \cite{kang2021modeling}. Several works have been carried out to promote the development of tendon-driven under-actuated robotic fingers. Xiong et al. \cite{7482814} proposed an anthropomorphic hand with innovative kinematic mechanisms and a limited number of actuators for replicating human grasping functions. Li et al. \cite{li2022brl} introduced the BRL/Pisa/IIT SoftHand—a low-cost, 3D-printed, underactuated, tendon-driven hand that leverages soft and adaptive synergies. Zhao et al. \cite{zhao2023adaptive} proposed a 3-DOF under-actuated finger based on a predefined elastic force gradient that decouples joints and enables conditional synergistic control, thereby significantly enhancing grasp adaptability and manipulative dexterity. However, these designs adopt relatively complex tendon routing, increasing manufacturing and assembly difficulty, and often do not achieve a substantial reduction in actuator count at the system level.


Researchers in the field have also attempted to enhance the stiffness and improve the grasping performance through various methods. Yan et al. \cite{9928329} proposed a C-shaped bidirectional stiffness joint for an anthropomorphic hand that enables low-stiffness tendon-driven bending and high-stiffness reverse bending. Legrand et al. \cite{10374190} designed a robotic anthropomorphic finger with variable stiffness achieved through multifunctional ligaments and miniature McKibben pneumatic artificial muscles. Other mechanisms, including low-melting-point alloys\cite{10197567}, an antagonistic twisted string actuator\cite{9969176}, and a cantilever structure\cite{LI2024105730} have also been investigated to change the stiffness of under-actuated tendon-driven robotic fingers. While effective in adjusting stiffness, these approaches typically require additional actuators, which increase weight and volume and also compromise their compactness. Developing a compact under-actuated scheme with simple tendon routing, predictable high stiffness, and the ability to drive an entire finger with a single actuator remains an open challenge.

To address these limitations, this work presents a novel under-actuated tendon-driven robotic finger (UTRF) that employs a simple yet effective synchronous tendon routing. This configuration ensures deterministic joint coupling, enabling the entire finger to be actuated by a single driver while achieving a balance between stiffness and compliance. The main contributions of this work are as follows:
\begin{itemize}
 \vspace{0mm}
    \item UTRF and its simple synchronous tendon routing are introduced in detail. The workspace of UTRF is calculated using the proposed kinematic model.
    \item The static model of UTRF is established, and the stiffness of the structure is analyzed, taking into account the elasticity of the tendons.
    \item The static loading experiment is conducted to validate the veracity of the static model with the fabricated prototype of UTRF.
    \item The prototype of UTRF-RoboHand based on the proposed UTRF is fabricated, and its mechanisms and electronics are introduced. Furthermore, the UTRF-RoboHand grasping experiments are conducted to test its compliance and dexterous manipulation capabilities.
\end{itemize}

\begin{figure*}[b]
	\begin{center}
		\includegraphics[width=5.5in]{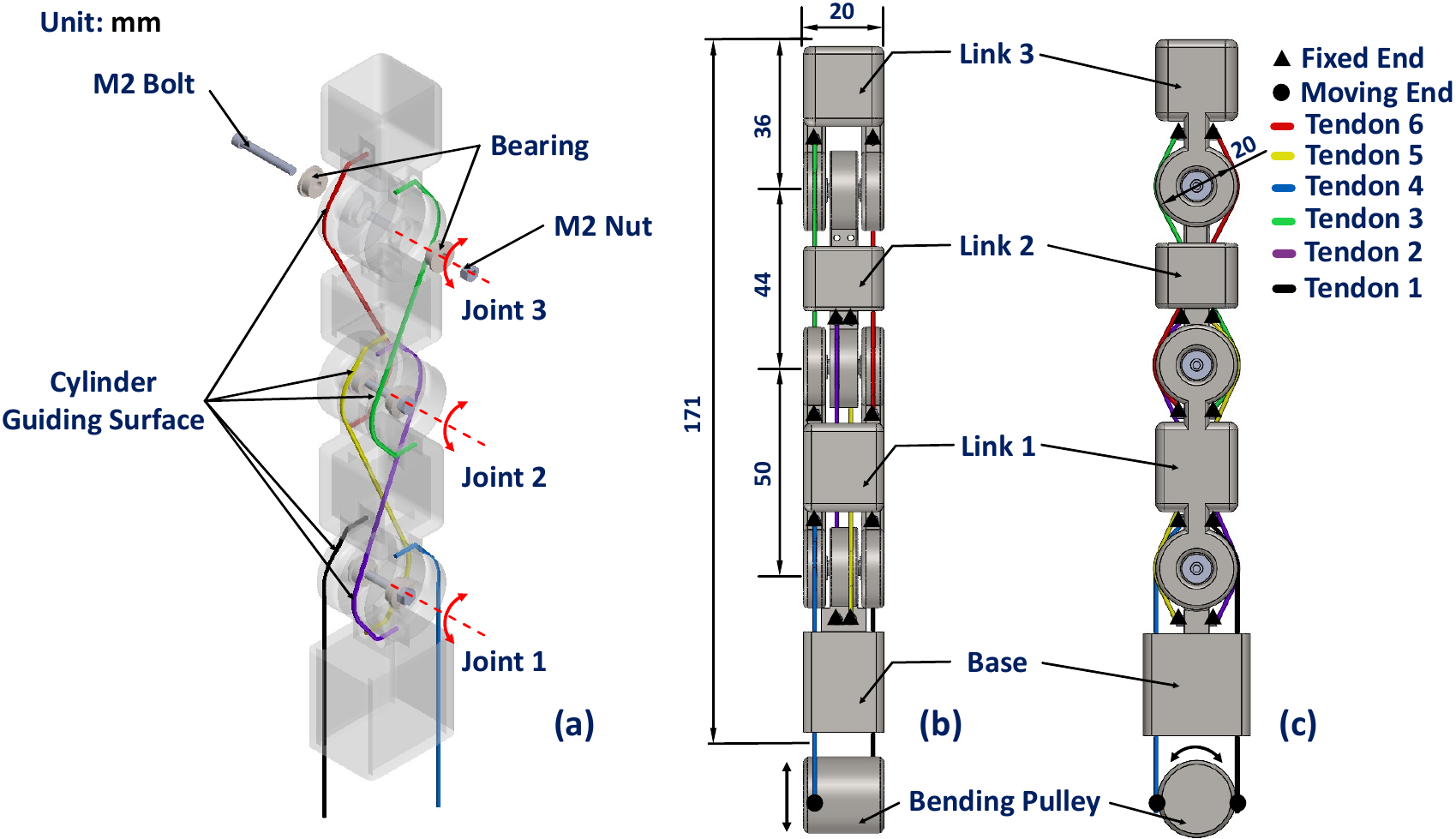}
		\caption{3D model of the designed UTRF. (a) Perspective view of UTRF with synchronous tendon routing, (b) Front view with tendons, (c) Left view with tendons.}
	\label{fig:1}
	\end{center}
\end{figure*}

\section{Kinematic modeling}
\subsection{Mechanism Design of UTRF}

The CAD model of the designed UTRF is illustrated in Fig. \ref{fig:1}. Exploded, front, and left views of UTRF are shown in Fig. \ref{fig:1}(a), Fig. \ref{fig:1}(b), and Fig. \ref{fig:1}(c), respectively. A simplified UTRF is comprised of three Links, a Base, coupling tendons (Tendon 2, 3, 5, and 6), actuating tendons (Tendon 1 and 4), and an actuating unit (Bending Pulley). The three Links and Base are connected end to end utilizing rotating hinges, thereby forming three-finger joints. Two actuating tendons are employed to connect Link 1 and the Bending Pulley in two directions. 

The synchronous tendon routing differentiates UTRF from conventional under-actuated designs. As shown in Fig. \ref{fig:1}(b–c), each coupling tendon connects two adjacent links and wraps around the cylinder guiding surfaces of both links in a precisely defined direction and wrap angle, with radii $R_1$, $R_2$, and $R_3$ corresponding to the guiding surfaces of Joint 1, Joint 2, and Joint 3, respectively. This routing geometry is intentionally designed so that when the proximal joint rotates, the coupling tendons enforce a fixed-length constraint, driving the distal joint to rotate synchronously. Specifically:
\begin{itemize}
    \item Between Joint 1 and Joint 2 (see Fig. \ref{fig:1}(b), Tendons 2 and 5), the coupling tendons wrap around the guiding surfaces of Base and Link 2. The wrap radius ratio $R_1/R_2$ determines the angular transmission ratio between the two joints.
    \item Between Joint 2 and Joint 3 (see Fig. \ref{fig:1}(b), Tendons 3 and 6), the same principle applies, with the wrap radius ratio $R_2/R_3$ defining the distal joint motion.
    
\end{itemize}
Because the tendons are grouped into Group A (Tendons 1–3) for flexion and Group B (Tendons 4–6) for extension, a single actuator pulling one actuating tendon can drive all three joints synchronously in either direction. This simple tendon winding configuration not only simplifies the mechanical structure but also provides both adequate stiffness for load-bearing and compliance for adaptive grasping, requiring only one actuator to control all joints of the finger.


\subsection{Kinematics of UTRF}

\begin{figure}[h]
	\begin{center}
		\includegraphics[width=2.9in]{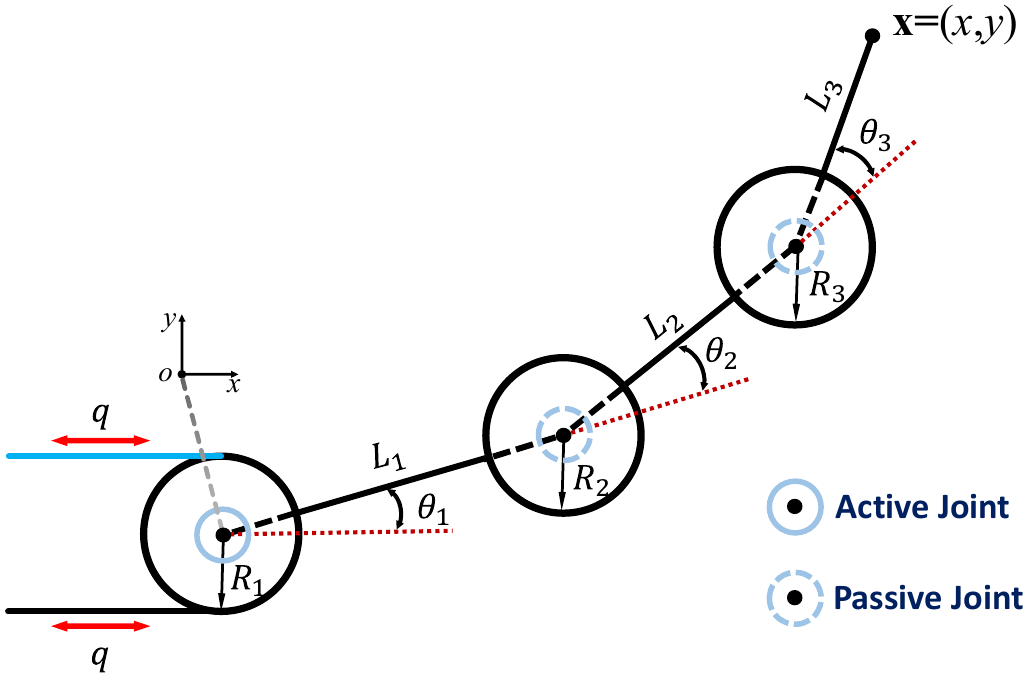}
		\vspace{-2mm}
		\caption{Simplified Kinematic Diagram of UTRF.}
		\label{fig:2}
	\end{center}
 \vspace{-6mm}
\end{figure}

The kinematic diagram of UTRF is shown in Fig. \ref{fig:2}. $\theta_i \ (i=1, 2, 3)$ are the rotation angles of Link $i$ ($i$=1, 2, 3) around Joint $i$ ($i=1,2,3$), respectively. It should be noted that the range of $\theta_1$ is $[ - {90^ \circ },{90^ \circ }]$. $L_i \ (i=1, 2, 3)$ is the length of Link $i$. $R_i \ (i=1, 2, 3)$ represents the radius of the $i$-th cylinder guiding surface. $q$ denotes the distance moved by Tendon 1 or 4. The position coordinate of the end of UTRF is $\textbf{x}=(x,y)$. It can also be described as
\begin{equation}
\left\{ {\begin{array}{*{20}{c}}
{x({\theta _i}) = \sum\limits_{i = 1}^i {{L_i}\cos \sum\limits_{i = 1}^i {{\theta _i}} } }\\
{y({\theta _i}) = \sum\limits_{i = 1}^i {{L_i}\sin \sum\limits_{i = 1}^i {{\theta _i}} } }
\end{array}} \right.\label{eq:1}
\end{equation}
where $\theta_i$ can be expressed as
\begin{equation}
{\theta _i} = {\frac{1}{{{R_i}}} \cdot } q\label{eq:2}
\end{equation}

Substituting Eq. (\ref{eq:2}) into Eq. (\ref{eq:1}), the following expression is obtained
\begin{equation}
\left\{ {\begin{array}{*{20}{c}}
{x(q) = \sum\limits_{i = 1}^i {{L_i}\cos \left( {\sum\limits_{i = 1}^i {\frac{1}{{{R_i}}} \cdot q} } \right)} }\\
{y(q) = \sum\limits_{i = 1}^i {{L_i}\sin \left( {\sum\limits_{i = 1}^i {\frac{1}{{{R_i}}} \cdot q} } \right)} }
\end{array}} \right.\label{eq:3}
\end{equation}

Then, the time partial derivative of Eq. (\ref{eq:3}) can be calculated and expressed in the matrix form
\begin{equation}
\left[ {\begin{array}{*{20}{c}}
{\dot x}\\
{\dot y}
\end{array}} \right] = \left[ {\begin{array}{*{20}{c}}
{ - \sum\limits_{i = 1}^i {\frac{{{L_i}}}{{{R_i}}}\sin \left( {\sum\limits_{i = 1}^i {\frac{1}{{{R_i}}} \cdot q} } \right)} }\\
{\sum\limits_{i = 1}^i {\frac{{{L_i}}}{{{R_i}}}\cos \left( {\sum\limits_{i = 1}^i {\frac{1}{{{R_i}}} \cdot q} } \right)} }
\end{array}} \right]\dot q\label{eq:4}
\end{equation}

Eq. (\ref{eq:4}) can be rewritten as
\begin{equation}
\bm{\dot x} = \bm{J} \cdot \dot q
\end{equation}
where $\bm{x}$ is the position vector of the end of UTRF, $\bm J$ is velocity Jacobian matrix of UTRF.

\subsection{Workspace of UTRF}

\begin{figure}[t]
	\begin{center}
		\includegraphics[width=3in]{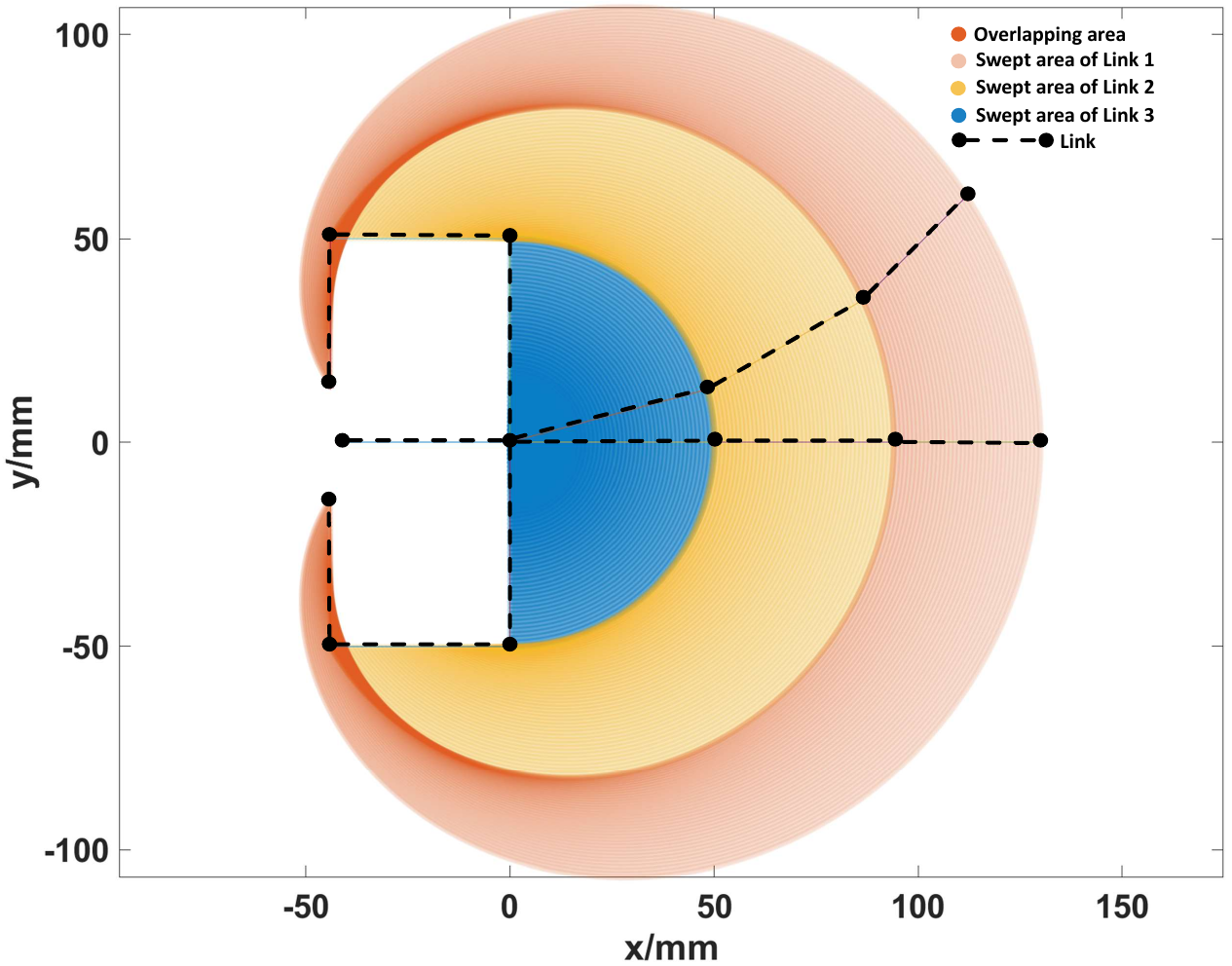}
		\caption{Workspace of UTRF.}
		\label{fig:3}
	\end{center}
 \vspace{-5mm}
\end{figure}

According to the kinematics analysis of UTRF, the workspace of UTRF can be calculated. Based on Eq. (\ref{eq:1}), points reached by Link $i$ can be calculated sequentially by iterating through $\theta$ and $L_i$. It should be noted that to obtain all the points swept by UTRF, L1, L2, and L3 are set as variables, with their range of values being from zero to their original values. The workspace of UTRF is shown in Fig. \ref{fig:3}. The blue, yellow, and red areas are the swept areas of Link 3, Link 2, and Link 1. The dark red area is the overlapping area between the areas swept by Link 1 and Link 2.

\normalem

\section{Statics analysis}
\label{sec:3}
\begin{figure}[h]
\centering
\includegraphics[width=3in]{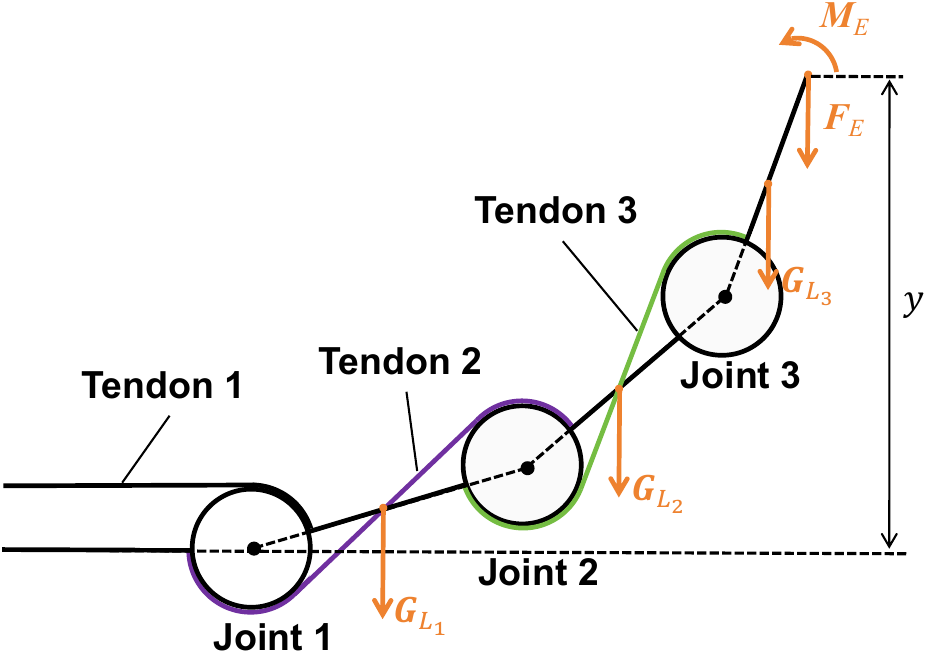}
\caption{The configuration of UTRF under external loading and prescribed actuating tendon displacement. The stretched tendons are noted, while slacked tendons are omitted here for better visualization. }
		\label{fig:4}
\vspace{-4mm}
\end{figure}

The static model is proposed to analyze the behavior of UTRF under external loading. Considering the typical manipulation condition, the configuration of UTRF in response to external loading and prescribed actuator tendon displacement can be illustrated in Fig. \ref{fig:4}. 

The external load consists of the gravitational force exerted by each finger link, the external force, and the moment acting on the fingertip. Typically, one set of the coupling tendons is slackened when the other set is stretched. To simplify the analysis process, the slack tendons are ignored while the stretched tendons are noted in the diagram. In accordance with the kinematic model, the configuration of the robotic finger is contingent upon the geometric constraints imposed by the actuating tendons and coupling tendons in terms of length. This can be determined by considering tendon tension and elasticity via the static model. 

\begin{figure*}[t]
\centering
\includegraphics[width=6.1in]{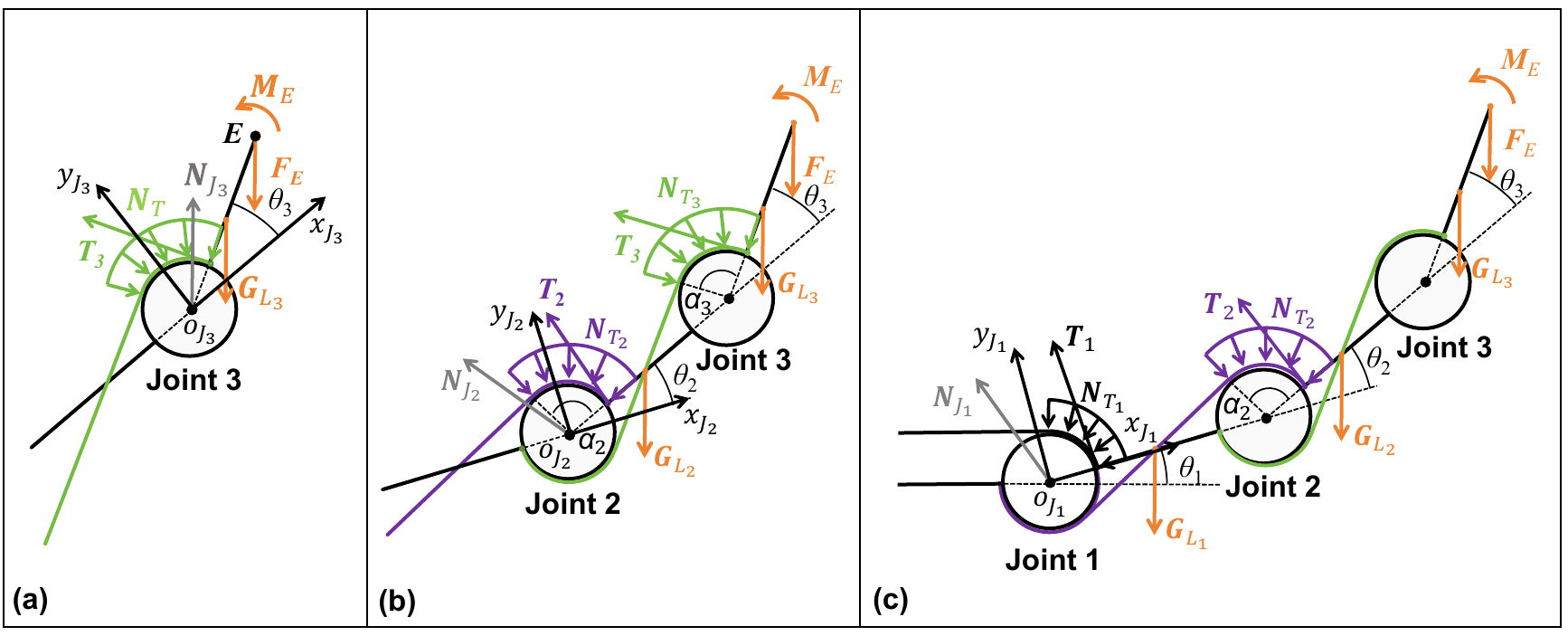}
\caption{Objects force analysis for each step, (a) Analysis of Link 3 in Step 1, (b)Analysis of the assembly of Link 3 and Link 2 in Step 2, (c) Analysis of the assembly of Link 3, Link 2, and Link 1 in Step 3. The detailed definitions of the symbols in the figure can be found in Appendix \ref{tab:1}.}
\label{fig:5}
\end{figure*}

Tendon tension in a given UTRF configuration is determined by analyzing the individual link assemblies, as illustrated in Fig. \ref{fig:5}. The detailed steps are as follows.

\subsubsection{Step 1}
Determine the tension in Tendon 3 by analyzing the force applied to Link 3. For a static stage, the equilibrium equation for the moments to $o_{J_3}$ is: 
\begin{equation}
\bm{M}_{\mathrm{E}}+\bm{M}_{o_{J_{3}}}^{F_{E}}+\bm{M}_{o_{J_{3}}}^{G_{L_3}}+\bm{M}_{o_{J_{3}}}^{T_{3}}=0    \label{eq:6}
\end{equation}
in which $\bm{M}_{o_{J_{3}}}^{F_{E}}$, $\bm{M}_{o_{J_{3}}}^{G_{L_3}}$, and $\bm{M}_{o_{J_{3}}}^{T_{3}}$ are the moments applied by $\bm{F}_{E}$, $\bm{G}_{L_3}$, and $\bm{T}_{3}$ to $o_{J_{3}}$ respectively. For a given finger configuration, $\bm{T}_{3}$ is the only unknown parameter in the above equation. Thus, the tension in Tendon 3 can be denoted using Eq. (\ref{eq:6}). 
\subsubsection{Step 2}
The same process can be used to determine the tension in Tendon 2 by analyzing the assembly of Link 2 and Link 3, which is shown in Fig. \ref{fig:5}(b). The equilibrium equation for the moments to $o_{J_2}$ can be denoted as:
\begin{equation}
\bm{M}_{\mathrm{E}}+\bm{M}_{o_{J_2}}^{F_{E}}+\bm{M}_{o_{J_2}}^{N_{T_3}}+\sum_{i=2}^{3}(\bm{M}_{o_{J_2}}^{G_{L_i}}+\bm{M}_{o_{J_2}}^{T_{i}})=0    \label{eq:7}
\end{equation}
in which $\bm{M}_{o_{J_2}}^{N_{T_3}}$ is the moment generated by $\bm{N}_{T_3}$, it can be calculated as:
\begin{equation}
\begin{split}
\bm{M}_{o_{J_2}}^{N_{T_3}}&=\int_{\theta_{3}}^{\alpha_{3}+\theta_{3}} \bm{N}_{T_3} \cdot L_{3} \cdot \sin (\theta) \cdot d \theta\\
&=\bm{N}_{T_3} \cdot L_{3} \cdot\left(\cos \left(\theta_{3}\right)-\cos \left(\alpha_{3}+\theta_{3}\right)\right)
\end{split}
\end{equation}
where $\alpha_{3}$ is the wrapping angle of Tendon 3 on the Cylinder Guiding Surface of Link 3. It can be calculated using the geometric relation:
\begin{equation}
\alpha_{3}=\pi-\arccos \left(\frac{R_{2}+R_{3}}{L_{2}}\right)-\theta_{3}  
\end{equation}
where $R_{2}$ and $R_{3}$ are the radii of the cylinder guiding surface of Link 2 and Link 3, respectively. Based on the above analyses, the tension of Tendon 2 can be acquired from Eq. (\ref{eq:7}). 
\subsubsection{Step 3}
Similarly, the assembly of Link 1, Link 2, and Link 3 can be employed to ascertain the tension value of Tendon 1, as illustrated in Fig. \ref{fig:5}(c). Tendon 3 is attached to Link 2 and Link 3 at two points. Therefore, the tensile force exerted by Tendon 3 does not contribute to the equilibrium of the assembly. The equilibrium equation for the moments to $o_{J_1}$ can be expressed as:
\begin{equation}
\begin{split}
&\bm{M}_{\mathrm{E}}+\bm{M}_{o_{J_1}}^{F_{E}}+\bm{M}_{o_{J_1}}^{N_{T_2}}+\sum_{i=1}^{3}\bm{M}_{o_{J_1}}^{G_{L_i}}+\sum_{i=1}^{2}\bm{M}_{o_{J_1}}^{T_{i}}=0
\end{split}\label{eq:10}
\end{equation}

Similarly $\bm{M}_{o_{J_1}}^{N_{T_2}}$ can be calculated as:
\begin{equation}
\begin{split}
\bm{M}_{o_{J_2}}^{N_{T_2}}&=\int_{\theta_{2}}^{\alpha_{2}+\theta_{2}} \bm{N}_{T_2} \cdot L_{2} \cdot \sin (\theta) \cdot d \theta\\
&=\bm{N}_{T_2} \cdot L_{2} \cdot\left(\cos \left(\theta_{2}\right)-\cos \left(\alpha_{2}+\theta_{2}\right)\right)
\end{split}
\end{equation}
where
\begin{equation}
    \alpha_{2}=\pi-\arccos \left(\frac{R_{1}+R_{2}}{L_{1}}\right)-\theta_{2}\label{eq:12}
\end{equation}

Thus the tension in all stretched tendons has been determined. For a stretchable tendon, the length of the tendon will increase due to elasticity. The stretched tendon length can be calculated using Hook’s Law:
\begin{equation}
    L_{T_\mathrm{i}}{ }^{\prime}=L_{T_i} \cdot(1+\frac{T_i}{E_i \cdot A_i}) (i=1,2,3)\label{eq:13}
\end{equation}
where $E_i$, $A_i$, and $L_{T_i}$ are Young’s modulus, cross-section area, and original length of Tendon $i$ respectively. The length of Tendon 1 is a set value, and the lengths of Tendon 2 and Tendon 3 can be calculated using geometric relations:
\begin{equation}
\begin{array}{l}
L_{T_2}=\left(\alpha_{20}-\cot \left(\alpha_{20}\right)\right) \cdot\left(R_{1}+R_{2}\right) \\
L_{T_3}=\left(\alpha_{30}-\cot \left(\alpha_{30}\right)\right) \cdot\left(R_{2}+R_{3}\right)
\end{array}  
\label{eq:14}
\end{equation}
where
\begin{equation}
\begin{array}{l}
\alpha_{20}=\pi-\arccos\left(\frac{R_{1}+R_{2}}{L_1}\right)\\
\alpha_{30}=\pi-\arccos\left(\frac{R_{2}+R_{3}}{L_2}\right)
\end{array}
\label{eq:15}
\end{equation}

Due to the elongation of the tendon length, the configuration of UTRF can be recalculated as:
\begin{equation}
\theta_{{i}}^{\prime}=\theta_{{i}}-\frac{L_{T_i}-L_{T_i}^{\prime}}{R_{i}}  
\label{eq:16}
\end{equation}

 The revised configuration will change the tendon tensions based on the equilibrium equations, and the adjusted tendon tensions will subsequently modify the position and orientation of UTRF. 
Algorithm \ref{algorithm2} is used to determine the final static configuration by analyzing the unloaded configuration along with the data on external loads. In the initial phase, the tensions of tendons are calculated based on the unloaded configuration. Subsequently, the configuration of UTRF is regenerated based on the tendon elasticity. Subsequently, the tendon tensions and finger configuration are recalculated recursively until the changes in the finger configuration between two adjacent loops satisfy $threshold$, which can be quantified by measuring the movement of the fingertip in the vertical direction.

\begin{algorithm}[]  
	\caption{Iterative Algorithm to Solve Static UTRF Configuration under External Loading and Prescribed Tendon Displacement.}
	\label{algorithm2}
	\LinesNumbered 
	\KwIn{${\rm{q}}, {E_i}, {A_i}, {R_i}, {L_i}, {\bm{G}_{Li}}, {\bm{F}_E}, {\bm{M}_E}, threshold$}
	\KwOut{$\theta _{{i}}^{\rm{k}},y^k,\bm{T}_{\rm{i}}^k$}
	\textbf{Initialize:} Set ${\rm{k}}$ to zero and calculate $\bm{L}_{T_i}^{\rm{k}}$ using geometric constrain Eq. (\ref{eq:14}), (\ref{eq:15})\; 
 
\Repeat{$\left| {y^k - y^{{\rm{k - 1}}}} \right| \le threshold$}
{ 
 $[\theta _{{i}}^{\rm{k}},y^k] \leftarrow $Solving kinematic Eq. (\ref{eq:2}) and Eq. (\ref{eq:16}) $[q,{R_i},{L_i},{L_{Ti}}]$\;
 $\bm{T}_{{i}}^{\rm{k}} \leftarrow $Solving equilibrium Eq. (\ref{eq:6}), (\ref{fig:7}), (\ref{eq:10}) $[{\bm{G}_{L{\rm{i}}}},{\theta _i},{\bm{F}_E},{\bm{M}_E}]$\;
 $L_{T{\rm{i}}}^{\rm{k}} \leftarrow $Elastic Eq. (\ref{eq:13}) $[{L_{T{\rm{i}}}},{E_i},{A_i},{\bm{T}_i}]$\;
 ${\rm{k}} \leftarrow {\rm{k}} + 1$\;}
 
\end{algorithm}

\section{Experiments}

In this section, prototypes of UTRF and UTRF-RoboHand are fabricated, followed by a static loading experiment of UTRF. The experiment aims to validate the static model presented in Section \ref{sec:3} and demonstrate the enhanced stiffness of the proposed mechanism. Furthermore, an experiment was conducted on UTRF-RoboHand to investigate the compliant and dexterous manipulation capabilities of UTRF.

\subsection{Static Loading Experiment}

\begin{figure}[t]
	\begin{center}
		\includegraphics[width=2.8in]{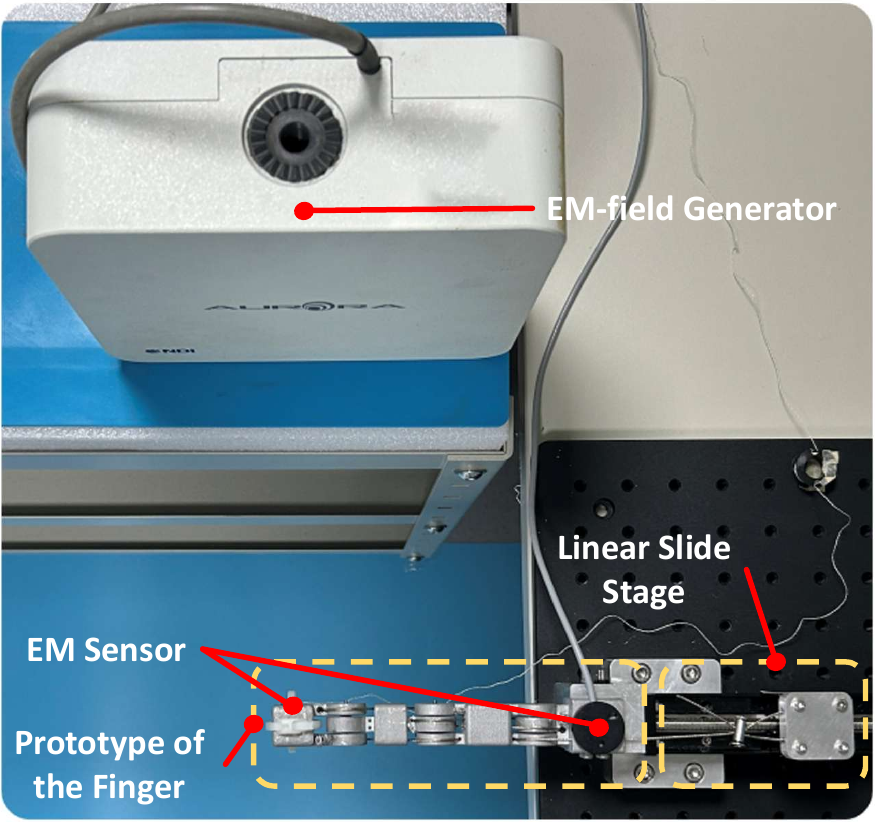}
		\caption{Setup of UTRF static loading experiment.}
		\label{fig:6}
	\end{center}
 \vspace{-7mm}
\end{figure}

The setup of the static loading experiment is illustrated in Fig. \ref{fig:6}. The prototype of UTRF has been manufactured using 3D printing technology based on the dimensions of the links as shown in Fig. \ref{fig:1}. All tendons are made of stainless steel with a diameter of 1 mm and Young's modulus of 200 GPa. The length of the actuating tendon has been set at 100 mm. An EM sensor is attached to the tip of UTRF, while an EM-field generator monitors its position and deflection. Prior to the commencement of the experiment, UTRF is driven to a horizontal, straight configuration using the Linear Slide Stage.

\begin{figure}[h]
	\begin{center}
		\includegraphics[width=2.8in]{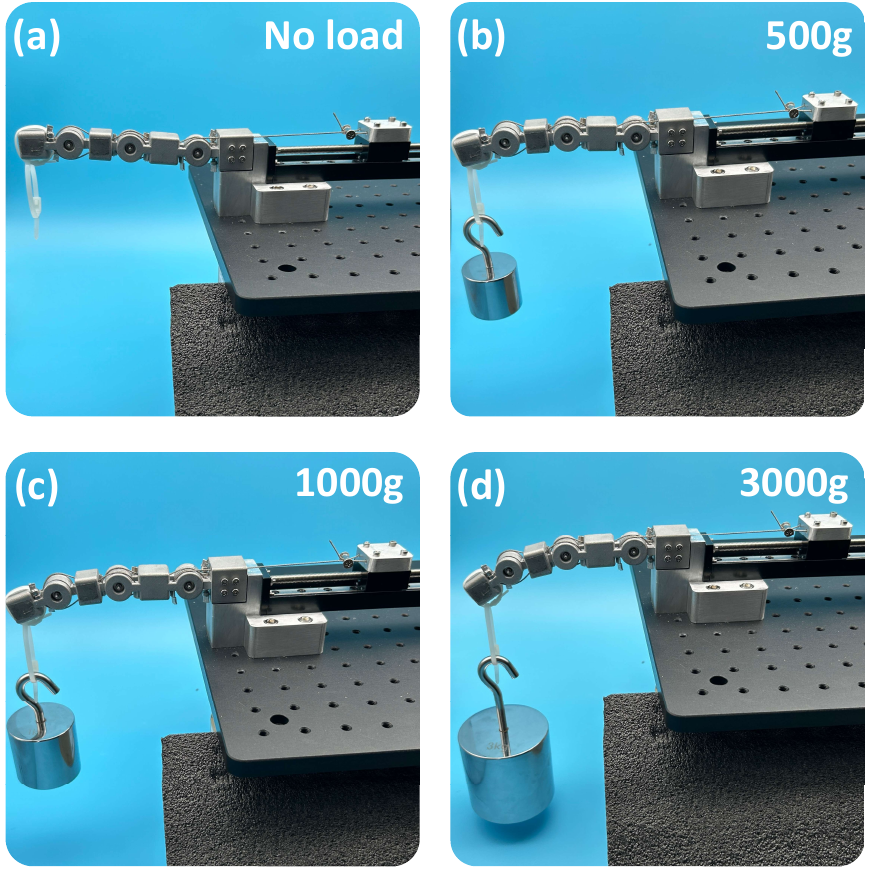}
		\caption{Deflection tests of UTRF with different payloads. (a) No load, (b) 500g, (c) 1000g, (d) 3000g.}
		\label{fig:7}
	\end{center}
    \vspace{-5mm}
\end{figure}

\begin{figure}[h]
    \begin{center}
        \includegraphics[width=2.9in]{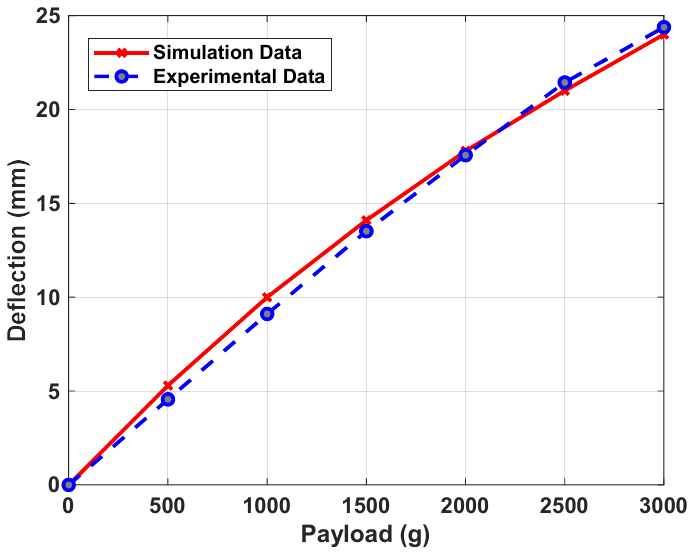}
	\caption{Static loading experiment results of UTRF prototype.}
	\label{fig:8}
    \end{center}
\end{figure}

In the experiment, the deflection of the UTRF prototype is measured under tip payloads of 500 g, 1000 g, 1500 g, 2000 g, 2500 g, and 3000 g, as shown in Fig. \ref{fig:7}. To eliminate the possibility of fortuitous outcomes, the loading experiment is conducted for 3 times each. The experimental results and the simulation data provided by the static model in Section \ref{sec:3} are presented in Fig. \ref{fig:8}. It can be observed that the simulation data are, for the most part, consistent with the experimental data. A comparison of the simulation data with the experimental data reveals that the maximum and mean data errors are 0.890 mm and 0.545 mm, respectively, which equates to 0.520 $\%$ and 0.322 $\%$ of the total finger length. The deflection is only 24.386 mm under an external load of 3000g, which is a stiffness of $1.2\times10^3 $ N/m, demonstrating the considerable stiffness and regulating ability of UTRF.


\subsection{Mechanisms and Electronics of UTRF-RoboHand}

\begin{figure}[h]
	\begin{center}
		\includegraphics[width=\columnwidth]{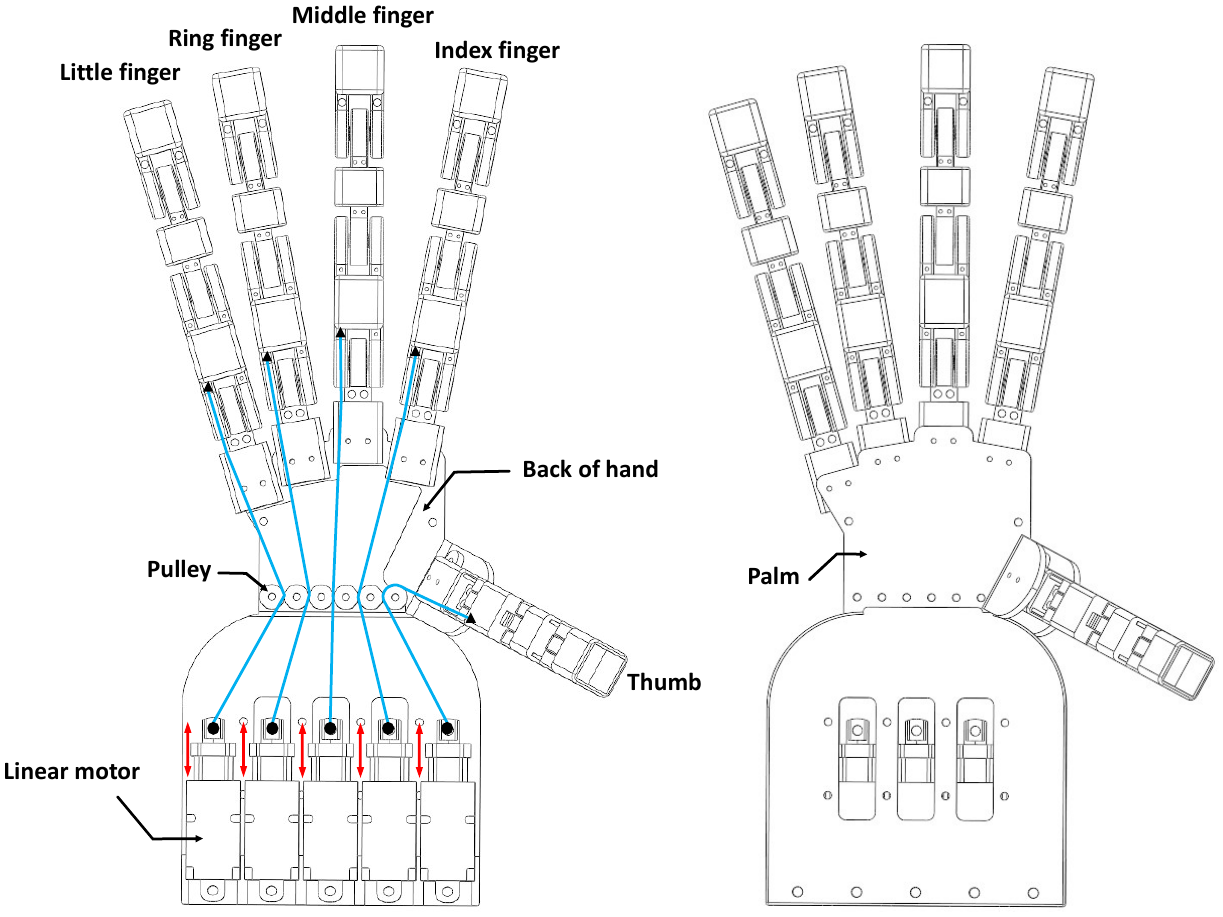}
		\caption{The mechanical structure of UTRF-RoboHand}
		\label{fig:9}
	\end{center}
\end{figure}

\begin{figure}[!h]
\vspace{-1mm}
    \begin{center}
            \includegraphics[width=3.0in]{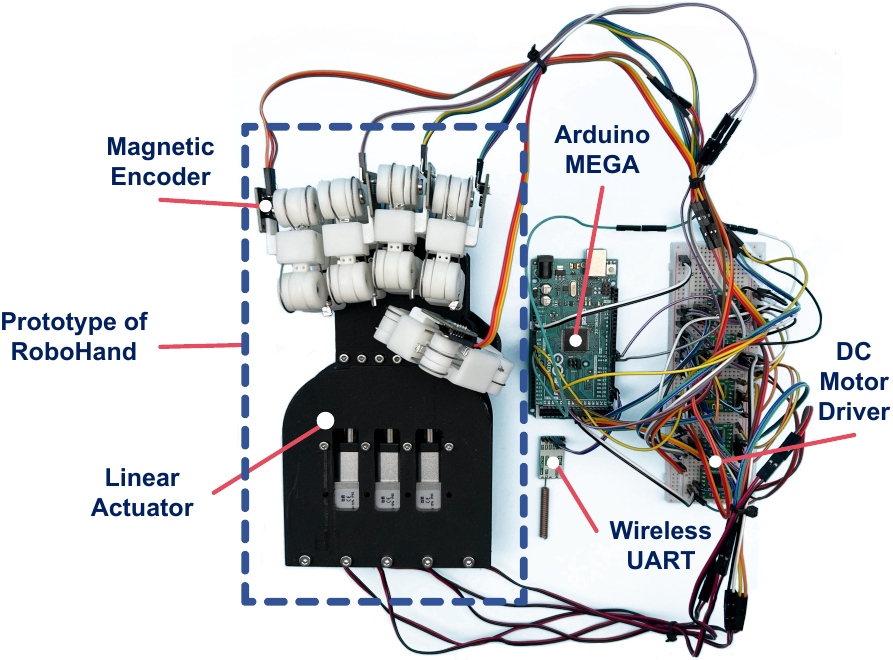}
	\caption{The prototype of UTRF-RoboHand with its electronic control design.}
    \label{fig:10}
    \end{center}
    \vspace{-5mm}
\end{figure}

In order to investigate the potential applications of UTRF, a prototype of UTRF-RoboHand is developed and utilized to conduct preliminary grasping tests. Fig. \ref{fig:9} illustrates the mechanical configuration of UTRF-RoboHand. It comprises five fingers, ten linear motors (two per finger; alternatively, five rotary motors, one per finger), ten pulleys, a palm, and a back-of-hand aspect. Each of the five fingers is constructed using UTRF. The grasping and releasing movements of each finger are actuated by two linear motors separately. One end of the tendon is secured to the moving end of the linear motor, passes through a pulley, and the other end is connected to Link 1 of the corresponding UTRF.

The prototype of UTRF-RoboHand is illustrated in Fig. \ref{fig:10}. The five fingers are produced using a resin-based printing material. The palm and the back of the hand are produced using PA6-CF. UTRF-RoboHand is designed with an integrated electronic control system, which enables precise and independent movement control for each finger. The controlling system is centered around an Arduino MEGA microcontroller. To drive the linear motors, the system employs TB6612 motor controller. One motor controller is capable of driving two linear motors with a rated load of 24 W. The control system uses 5 motor controllers to control 5 linear motors for system redundancy. A magnetic encoder is mounted on Joint 2 of each finger to achieve the closed-loop feedback control.

\subsection{UTRF-RoboHand Grasping Experiment}
\begin{figure}[t]
	\begin{center}
		\includegraphics[width=\columnwidth]{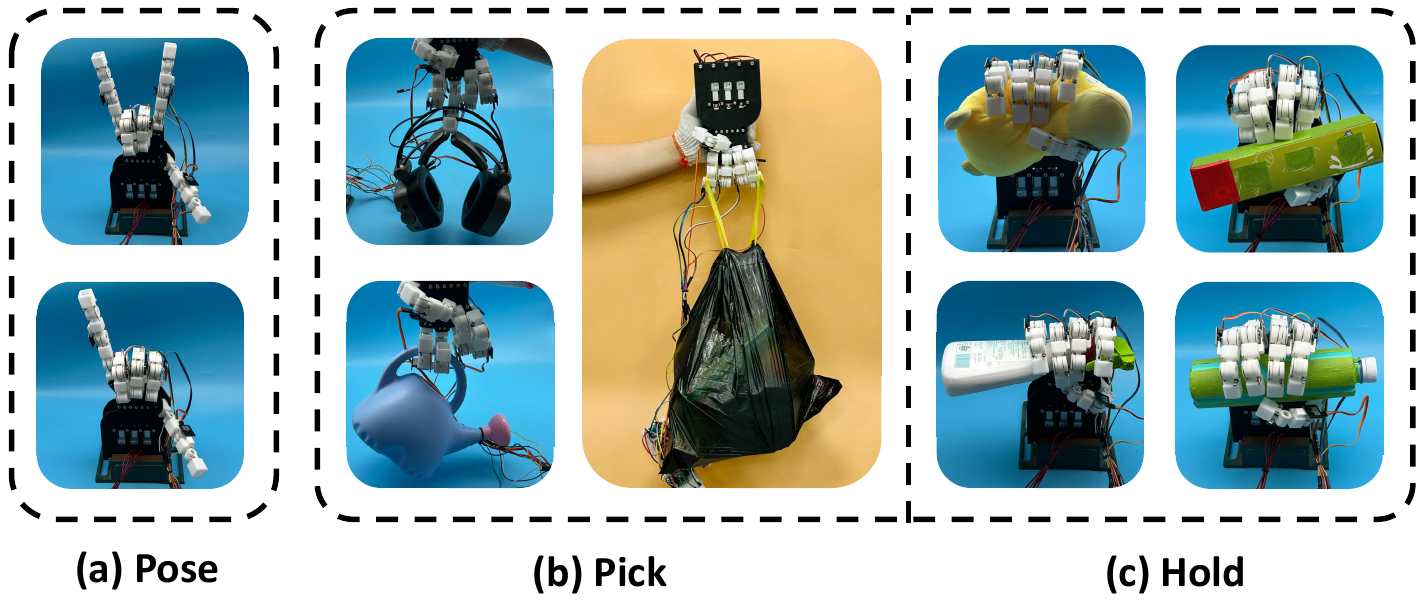}
		\caption{Different poses, picking, and grasping scenarios of the developed UTRF-RoboHand. (a) Pose, (b) Pick, and (c) Hold.}
		\label{fig:11}
	\end{center}
 \vspace{-2mm}
\end{figure}

In this experiment, the pose and grasping tests of UTRF-RoboHand are conducted to assess its manipulating capability. UTRF-RoboHand is operated to perform two poses: ``Rock and Roll'' (see Fig. \ref{fig:11}(a) above) and ``Shaka Sign'' (see Fig. \ref{fig:11}(a) below). Moreover, UTRF-RoboHand is programmed to perform a variety of poses, enabling it to grasp objects of different shapes, weights, and fragility in various scenarios, as shown in Fig. \ref{fig:11}(b). For each grasping test, the robotic hand is commanded to hold the object for 10 minutes and is disturbed by the hitting of the hammer. This demonstrates its strong compliance and manipulation capabilities.

\section{CONCLUSIONS}

This paper presents a novel under-actuated tendon-driven robotic finger to address the challenge of stiffness improvement by using a serial synchronous transmission mechanism for tendon actuation. 
The corresponding kinematic and static models, along with an accurate stiffness analysis, were derived. A single-finger prototype was specifically built to test and demonstrate the high stiffness of the proposed structure design, achieving a high stiffness of $1.2\times10^3 $ N/m, and to validate the deduced static model. 
The compliance and dexterity of the structure were further demonstrated by a full-scale five-finger robotic hand prototype in multi-object grasping and posing experiments. 
With its simple synchronous tendon routing that balances stiffness and compliance and enables single-actuator control of the entire finger, the proposed structure offers a promising approach for enhancing under-actuated tendon-driven robotic fingers, providing high load capacity and configurable stiffness supported by a reliable static model.

Future work may focus on further optimizing the design for specific applications and improving the control algorithms for more complex manipulation scenarios.




\appendix  
\setcounter{table}{0}   
\setcounter{figure}{0}
\setcounter{section}{0}
\setcounter{equation}{0}
\renewcommand{\thetable}{A\arabic{table}}
\renewcommand{\thefigure}{A\arabic{figure}}
\renewcommand{\thesection}{A\arabic{section}}
\renewcommand{\theequation}{A\arabic{equation}}
\begin{table}[h]
\setlength{\belowcaptionskip}{-0.1cm}
\begin{center}
\caption{The specification list of parameters in Fig. \ref{fig:5}}
\label{tab:1}
	\begin{tabular}{c l}
			SYMBOL & NOMENCLATURE\\
   &\\
			$o_{J_i}$ &  Original point of Joint $i$   \\
			
			$x_{J_i}$  &$x$-axis of Joint $i$\\
			
			$y_{J_i}$ & $y$-axis of Joint $i$\\
		
			$\bm{N}_{J_i}$  &Supporting force of the bearing on Joint $i$ \\
         $\bm{N}_{T_i}$ & Normal force exerted by Tendon $i$ on\\
               & Cylinder Guiding Surface.\\
               $\bm{T}_{i}$& Tensile force of Tendon $i$ at the \\
                & connection point of Link $i$.\\
                  $\bm{G}_{L_i}$ &  Gravity of Link $i$\\
                   $\theta_i$     & Absolute angle of Link $i$ rotating \\
                          & around Joint $i$\\
                        $\alpha_i$  & Wrapping angle of Tendon $i$ around \\
                        & Cylinder Guiding Surface of Joint $i$.\\
                      $\bm{F}_E$  & External force applied to point E.\\
                      & with an arbitrary direction.\\
                      $\bm{M}_E$ & External moment applied to point E.\\
                      &
                   
		\end{tabular}
\end{center}
\vspace{-6mm}
\end{table}



\bibliographystyle{IEEEtran}
\normalem
\bibliography{IEEEabrv, References}

\end{document}